\documentclass[12pt]{l4dc2021}

\title[]{Forced Variational Integrator Networks for Prediction and Control of Mechanical Systems}
\usepackage{times}
\usepackage{caption}
\usepackage{wrapfig}
\usepackage{algorithm, algorithmic}

\author{%
 \Name{Aaron Havens} \Email{ahavens2@illinois.edu}\\
 \Name{Girish Chowdhary} \Email{girishc@illinois.edu}\\
 \addr Department of ECE \& Coordinated Science Laboratory\\
University of Illinois at Urbana-Champaign, Champaign, IL, USA, 61820%
}

\begin{document}

\maketitle

\begin{abstract}
As deep learning becomes more prevalent for prediction and control of real physical systems, it is important that these overparameterized models are consistent with physically plausible dynamics. This elicits a problem with how much inductive bias to impose on the model through known physical parameters and principles to reduce complexity of the learning problem to give us more reliable predictions. Recent work employs discrete variational integrators parameterized as a neural network architecture to learn conservative Lagrangian systems. The learned model captures and enforces global energy preserving properties of the system from very few trajectories. However, most real systems are inherently non-conservative and, in practice, we would also like to apply actuation. In this paper we extend this paradigm to account for general forcing (e.g. control input and damping) via discrete d'Alembert's principle which may ultimately be used for control applications. We show that this forced variational integrator networks (FVIN) architecture allows us to accurately account for energy dissipation and external forcing while still capturing the true underlying energy-based passive dynamics. We show that in application this can result in highly-data efficient model-based control and can predict on real non-conservative systems.
\end{abstract}

\begin{keywords}
Variational Integrators, Physics-constrained learning, Model-based Control
\end{keywords}

\section{Introduction}
Deep neural networks (DNN) have seen increasing success in scaling and generalizing to large data sets for classification, prediction and even control~(\cite{imagenet, Silver1140}). However when it comes to physic systems like autonomous robots, generalization is especially important so that interpolations/extrapolations of training data still results in physically plausible and safe behavior~(\cite{guiochet2017safety}). In these physical applications we often choose to engineer carefully parameterized physics-based models over deep overparametrized models to ensure this level of generalization. These expertly engineered models impose a great amount of inductive bias (fundamental assumptions about the output function), but as we know incorrect bias can be detrimental to decision-making agents and fully modeling every contributing process can be costly and laborious~(\cite{dullerud2013course, model_bias}). Clearly these systems all share similar physical principles such as conservation laws and symmetries. How can we use highly flexible deep architectures and impose inductive bias in a way that generally applies to a large number of physical systems? Better yet, can they be interpretable through existing physical theory?

This is a challenging problem since there are real conserved and invariant quantities which make the difference between a system being stable or unstable and have major consequences in both long-term prediction and control. \citet{hamiltonianNN} highlights these issues and provides a method of directly parameterizing a system Hamiltonian with a DNN and then reconstructing the dynamics with classical mechanics techniques. 

On the other hand \citet{neural_ode} interprets deep residual networks as Euler discretized continuous time ordinary differential equations. \citet{pmlr-v108-saemundsson20a} provides a natural link from this perspective to a class of discretization techniques from discrete mechanics called Variational Integrators (VI)~(\citet{west}). Variational integrators formally discretize the action of a Lagrangian system, rather than the derivative, which approximately preserves energy and momentum (otherwise known as a sympletic integrator). \citet{pmlr-v108-saemundsson20a} uses an explicit variational integrator structure to propose VI networks (VIN) that can learn dynamics of conservative Lagrangian systems with very accurate energy behavior. The resulting inductive bias from this VI discretization yields data-efficient generalization for a large class of systems. However, none of the aforementioned methods take into account external forcing and non-conservative systems which are abundant in the physical world. \citet{LutterRP19} proposes an architecture that incorporates forces by learning the inertial matrix and the potential function by minimizing violations of Lagrangian dynamics. However the proposed loss function does not account for passive energy dissipation and doesn't explicitly enforce any kind of learned conservation of preserved quantities. Perhaps most related to our work, \cite{hochlehnert2021learning} concurrently proposed methods for learning conservative VIN's with discontinuous contact dynamics which could be complementary to this work.

\paragraph{Contributions}
In this paper we propose a direct extension of \citet{pmlr-v108-saemundsson20a} by applying the forced discrete variational principle \textit{Discrete Lagrange-d'Alembert Principle}~(\cite{junge2005discrete,west}). With this extension we are able to introduce general non-conservative forcing into an explicit VI that yields a Forced Variational Integrator Network (FVIN) learning architecture. FVIN can handle control inputs and dissipation while still enjoying the long term energy tracking properties of the original VIN, enabling a much wider scope of applications in control. We also highlight the efficient generalization properties of FVIN in low data regimes as few as $5$ trajectories on two nonlinear nonconservative systems. We further demonstrate the importance of this extension by applying FVIN to predict a real non-conservative cart-pole system from data.
\section{Preliminaries}
We are motivated to introduce a class of integrators which define a discrete-time dynamical system which preserves global energy and momentum properties of the corresponding continuous system. A brief background of Lagrangian mechanics with forcing will be presented along with their lesser-known discrete analogs in the form of variational integrators. Specifically, we look to control mechanical systems whose Lagrangian $L:TQ \rightarrow \mathbb{R}$ is given as the difference of kinetic and potential energies.
\begin{equation}
\label{eq:lagrangian}
L (q,\dot q) = T(\dot q) - V(q)  = \frac{1}{2}\dot q^{\intercal}M \dot q - V(q)
\end{equation}
Here we have a configuration state $q$ belonging to $Q$ which, in general, could be an $n$-dimensional configuration manifold. However in this paper we restrict our focus to integrators on the euclidean space $\mathbb{R}^n$. The state space is then given by the tangent bundle $TQ$ where at a point $q$ we have position and velocity $(q,\dot q)$. Using this interpretation the state space $TQ$ is the euclidean space $\mathbb{R}^{2n}$. $M$ is a symmetric positive definite mass matrix and $V(q)$ is the energy potential function. We also consider external forcing to account for actuators and passive dissipation with the Lagrangian force $f_L: T Q \rightarrow T^* Q$, where $T^* Q$ is the cotangent bundle. We can obtain the governing dynamics by finding extremizing curves in $\mathcal{C}(Q)$ that minimize an action functional, where $\mathcal{C}(Q) = \{q:[0,T] \rightarrow Q \mid q \in \mathcal{C}^2 \}$ is the path space (essentially twice differentiable curves). These extreme curves can be characterized by \textit{Lagrange-d'Alembert principle}~(\cite{vujanovic1978conservation}) and satisfy
\begin{equation}
    \label{eq:action}
\delta \int^T_0 L(q(t), \dot q(t))dt + \int^T_0 f_L(q(t),\dot q(t), u(t)) \cdot \delta q(t) dt = 0
\end{equation}
where $\delta$ are variations of $\mathcal{C}(Q)$ such that $\delta q(0) = \delta q(T) = 0$.
This necessary condition gives us a description of the configuration dynamics resembling a forced variant of the standard Euler-Lagrange equations. 
\begin{equation}
\label{eq:dalembert}
    \frac{\partial L}{\partial q}(q,\dot q) - \frac{d}{dt}\bigg(\frac{\partial L(q,\dot q)}{\partial \dot q}\bigg) + f_L(q,\dot q, u) = 0
\end{equation}
At this point, one could discretize Eq~\ref{eq:dalembert} directly to obtain an explicit forward Euler scheme with time step $h$. This could be interpreted as a residual dynamic network update rule.
% \begin{equation*}
% M\Ddot{q} +\nabla{V(q)} =0,\quad
% \begin{pmatrix} q_{k+1}\\ \dot q_{k+1} \end{pmatrix} =\begin{pmatrix} q_{k}\\ \dot q_{k} \end{pmatrix} + h\begin{pmatrix} \dot q_k \\ -M^{-1}\nabla{V(q_k)} \end{pmatrix}
% \end{equation*}
But then what can we say about aforementioned energy conservation properties? Since discretization was performed directly on the resulting ODE, in some sense we lose information given by the Lagrangian. We can instead apply discretization to the integral Eq.~\ref{eq:action} directly and derive a new integration rule which will more accurately conserve total energy under the action (sometimes called sympletic integrators). This is the purpose of \textit{Variational Integrators}~(\cite{wendlandt1997mechanical,west}) which define implicit and explicit discrete update rules using a discrete analog to the Lagrange-d'Alembert principle.

\subsection{Variational Integrators}
Variational integrators are aimed to provide a discrete analog to the continuous time Euler-Lagrange principle, which involves taking an approximation of the action integral. This idea is introduced by the \textit{Discrete Lagrangian} defined by a quadrature-based approximation of the action integral found in Eq.~\ref{eq:action}.
\begin{equation*}
\label{eq:discrete_l}
L_d(q_k, q_{k+1}) \approx \int_{t_k}^{t_{k+1}} L(q(t),\dot q(t)) dt
\end{equation*}
We approximate the Lagrangian forcing by defining left $f^-_d(q_k,q_{k+1},u_k)$ and right $f^+_d(q_k, q_{k+1}, u_k)$ \textit{discrete Lagrangian forces} which are then used to approximate the variation of the continuous external force in Eq.~\ref{eq:action} over a small interval:
\begin{equation*}
    \label{discrete_action}
    f^-_d(q_k,q_{k+1},u_k)\cdot \delta q_k + f^+_d(q_k,q_{k+1},u_k)\cdot \delta q_{k+1} \approx \int^{t_{k+1}}_{t_k} f(q(t), \dot q (t), u(t)) \cdot \delta q dt
\end{equation*}
Using these approximations with a similar variational argument yields a discrete analog to 
%variation in terms of a sum between two boundary points $q_0$ and $q_N$. Intuitively we have defined the the integral by $N$ points on the curve sampled uniformly in time. These discrete curves $\{q_k \}^N_{k=0}$ satisfy
% \begin{equation}
%     \delta \sum^{N-1}_{k=0} L_d(q_{k},q_{k+1}) + \sum^{N-1}_{k=0} [ f^-_d(q_k,q_{k+1},u_k)\cdot \delta q_k + f^+_d(q_k,q_{k+1},u_k)\cdot \delta q_k]=0
% \end{equation}
% for all variations $\{\delta q_k \}^N_{k=0}$ such that $\delta q_0 = \delta q_N = 0$.
%Applying a variational argument to our action sum will yield the discrete form of
our forced Euler-Lagrange equations. For each $k$ the following must hold, defining an integrator implicitly. Here $D_i$ denotes the slot derivative differentiating with respect to the $i$-th argument
\begin{equation}
    \label{eq:position_int}
    D_2 L_d(q_{k-1},q_k) + D_1 L_d(q_{k}, q_{k+1}) + f^+_d(q_{k-1},q_k, u_k) + f^-_d(q_k, q_{k+1}, u_k)= 0
\end{equation}
In view of the discrete Legendre Transform, described in great detail in~\cite{west}, we can obtain an equivalent position-momentum form corresponding to the discrete Hamiltonian where the momentum for the Newtonian system is given by $p= \frac{\partial L}{\partial \dot q} = M \dot q$.
\begin{equation}
\label{eq:velocity_int}
\begin{aligned}
    p_k = - D_1 L_d(q_{k},q_{k+1}) - f^-_d(q_k, q_{k+1}, u_k),\quad
    p_{k+1} = D_2 L_d(q_{k}, q_{k+1}) + f^+_d(q_k, q_{k+1}, u_k)
\end{aligned}
\end{equation}
Given the Lagrangian in the form of Eq.~\ref{eq:lagrangian}, we can actually write both integrators~(\ref{eq:position_int}, ~\ref{eq:velocity_int}) explicitly as a function of $(q_k, q_{k-1}, u_k)$ and $(q_k, \dot q_k, u_k)$ respectively. These explicit integrators  will be introduced in the following section.
\subsection{Explicit Integrators with Forcing}
The integrators that will be ultimately useful in designing a feed-forward network architecture will be those which involve no numerical scheme and give the next state explicitly in terms of previous states and controls. To obtain this explicit integrator we take advantage of our chosen approximation to the discrete Lagrangian. We can use a symmetric quadrature rule to approximate the discrete Lagrangian to derive both the position and position-momentum integrators explicitly:
\begin{equation}
    \label{eq:quadrature}
     L_d(q_k, q_{k+1}) = \frac{h}{2}\bigg( L\left(q_k, \frac{q_{k+1} - q_k}{h}\right) + L\left(q_{k+1}, \frac{q_{k+1}-q_k}{h}\right) \bigg)
\end{equation}
Here we refer to the Lagrangian given in Eq.~\ref{eq:lagrangian} and a uniform time-step $h$ although in principle it can vary. For the position integrator, we choose a simplified form of the discrete forces under a \textit{zero-order hold} assumption distributed only over $f^+_d$, where $F(q_k, u_k)$ maps control inputs approximately to discrete Lagrangian forces
\begin{align*}
    f^-_d(q_{k},q_{k+1}, u_k) &= h F(q_k, u_k)\\
    f^+_d(q_{k}, q_{k+1}, u_k) &= 0
\end{align*}
This choice in particular will give us an explicit position integrator akin to the \textbf{\textit{Stormer-Verlet}} (SV) integrator~\citet{verlet}
\begin{equation}
\label{eq:stormer}
    q_{k+1}= 2 q_k - q_{k-1} + h^2 M^{-1}(F(q_k, u_k) - \nabla V(q_k))
\end{equation}
For our position-velocity integrator we choose a symmetric representation of the discrete forces which are distributed evenly over both $f^+_d$ and $f^-_d$
%as $f^\pm_d(q_k, q_{k+1}, u_k) = \frac{h}{2} F(q_k, u_k)$.
\begin{equation*}
    f^\pm_d(q_k, q_{k+1}, u_k) = \frac{h}{2} F(q_k, u_k)
\end{equation*}
This yields an explicit position-velocity integrator otherwise known as \textbf{\textit{Velocity-Verlet}} (VV) integrator. Note again for our Newtonian system that momentum and velocity relate by $p = M \dot q$
\begin{equation}
\label{eq:verlet}
    \begin{aligned}
        q_{k+1} &= q_k + h \dot q + \frac{h^2}{2}M^{-1}(F(q_k, u_k) - \nabla V(q_k))\\
        \dot q_{k+1} &= \dot q_k + h M^{-1}\bigg(F(q_k, u_k) - \frac{\nabla V(q_k) + \nabla V(q_{k+1})}{2}\bigg)
    \end{aligned}
\end{equation}
Now that we have an explicit expression for the state evolution, we can inspire a feed-forward architecture where functions $V$ and $F$ be learned directly from data from possibly non-conservative systems. Focusing on the VV integrator, we show that this structure can effectively capture these terms in physically meaningful ways, being able to precisely account for multiple types of forcing even in very low data regimes.  
\section{Forced Variational Integrator Networks}
\begin{figure}
    \centering
    \includegraphics[width=0.75\textwidth]{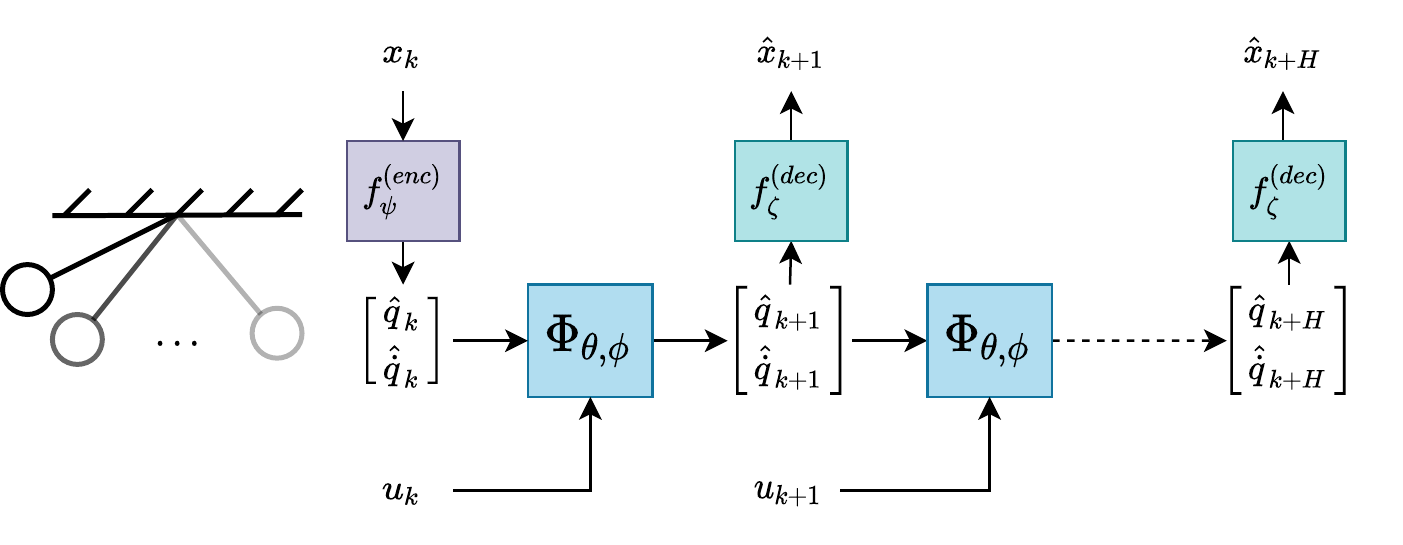}
    \caption{The FVIN predicts ahead multiple steps from an initial condition and a sequence of controls in an open-loop fashion. The loss is then computed against these roll-outs to encourage long-term consistency.}
    \label{fig:my_label1}
\end{figure}
With an explicit integrator that supports forcing in hand, we may now discuss how this structure can be used to design a general purpose neural network predictor for non-conservative Newtonian systems. Enforcing that a feed-forward neural network takes on the form of a variational integrator lets us arrive at predictors with correct global energy and interpretable forcing (e.g. damping, actuation) with much fewer samples. As a result, such architectures (like the integrators they are inspired by) have better long term prediction and and can be used to implement higher performing model-based controllers for a broad class of systems. We begin by considering the case where configuration states are measured directly. The SV~(\ref{eq:stormer}) integrator takes the argument $(q_k,q_{k-1})$ where VV~(\ref{eq:verlet}) takes in $(q_k, \dot q_k)$. The potential function and forcing function are parameterized by $V_{\theta}$ and $F_{\phi}$ as simple feed-forward neural networks. We choose to absorb the mass matrix $M$ into these learned objects, although, in principle, it may be learned as well. Although one can differentiate $V$ we found that estimating $M^{-1}\nabla V_{\theta}$ directly makes no difference in practice and is more efficient to evaluate in the forward pass. 
\begin{align}
V_{\theta}(q_k) \approx M^{-1} V(q_k),\quad
F_{\phi}(x_k, u_k) \approx M^{-1}F(q_k, u_k)
\end{align}
\subsection{Encoding Observations into Configurations}
Oftentimes, a proper configuration state cannot be observed directly and the state must be estimated from observations. We may do this using a type of dynamic autoencoder scheme (related to~\cite{kingma2013auto})), where the VIN dynamics are applied to encoded states before being decoded. Note that one could use a variational autoencoder scheme as done by~\cite{pmlr-v108-saemundsson20a}, however we focus on the deterministic setting for exposition. We can learn all of these functions simultaneously simply by minimizing the multi-step ``open-loop'' loss in Eq.~\ref{eq:loss} where we decode a sequence of configuration states rolled out from the initial encoded state. The encoder and decoder networks are denoted by $f^{(enc)}_{\psi}$ and $f^{(dec)}_{\zeta}$ respectively and we aggregate parameters under $\Theta = (\theta, \phi, \psi, \zeta)$. The loss is computed under the squared $l_2$ loss up to time $T$. We denote both SV and VV integrators using $\Phi_{\theta,\phi}$.
\begin{equation}
\label{eq:loss}
\begin{aligned}
    &&\underset{\Theta}{\text{minimize}}\quad &\sum^{T-1}_{k=0} ||\hat y_{k+1}- y_{k+1}||_2^2\\
    &&\text{subject to} \quad \hat x_0 &= f^{(enc)}_{\psi}(y_0)\\
    &&                \hat y_k &= f^{(dec)}_{\zeta}(\hat x_k)\\
    &&               \hat x_{k+1} &= \Phi_{\theta,\phi}(\hat x_{k}, u_{k}) \quad \forall k \in \{0,\ldots, T-1\}
\end{aligned}
\end{equation}
where here our labels $y_k$ are simply the true next configuration states, $y_k=(q_k)$ for SV and $y_k = (q_k, \dot q_k)$ for VV integrators. In the case that configuration come directly as observations, the encoding and decoding networks are simply identity operators.
\subsection{Inductive Bias in Forcing}
Forcing can be introduced in several ways, but most commonly in mechanical systems forcing is comes in the form of actuation, damping and heat. In some cases, we have additional information such that damping is only a function of velocity or that forcing is independent of state entirely. In either of these cases, we can decouple the parameters $\phi$ into $\phi_1$ and $\phi_2$ for control and damping respectively. For the general case going forward we will use the following structure for the VV-FVIN forces
\begin{equation}
    F_{\phi_{1,2}}(q_k, \dot q_k, u_k) = F_{\phi_1}^{(control)}(q_k, u_k)+F_{\phi_2}^{(damping)}(q_k, \dot q_k)
\end{equation}
For the SV-FVIN we chose to capture velocity information in damping using the current and previous states where control remains consistent with the VV-FVIN.
\begin{equation}
    F_{\phi_{1,2}}(q_{k-1}, q_k, u_k) = F_{\phi_1}^{(control)}(q_k, u_k)+F_{\phi_2}^{(damping)}(q_{k-1},q_k)
\end{equation}
%===============================================================================
\section{Experimental Results}
In this section we will investigate several aspects of performance for the proposed FVIN, specifically the VV-FVIN variant (performance results for SV-FVIN prediction can be found in Appendix~\ref{app:sv}). The prediction and control performance will be evaluated under two forms of forcing mentioned previously, actuation and damping. We will also consider generalization characteristics such as the FVIN's ability to capture the true energy conserving behavior in the absence of forcing. This would show that the model is able to truly capture the properties that we claim to describe in the FVIN, rather than simply being a black-box residual model. We study the damped pendulum ($1$D rotational configuration state), the damped cartpole ($2$D configuration state with rotation and translation) and a real rotational cartpole system. Although these systems may seem simplistic, they are highly-nonlinear and canonical examples of real mechanical systems, yet are easy to interpret.
\subsection{Baseline Residual Dynamic Network}
To evaluate both prediction and control of the proposed FVIN framework, we compare against a similar architecture where the FVIN dynamics are replaced by a general residual dynamic network (ResNN) as in~\cite{pmlr-v108-saemundsson20a}, trained over the $H$-step loss function in Eq.~\ref{eq:loss}. Additionally we give the residual network the same additive forcing assumption as the FVIN so that, with encoding and decoding network structure being held constant, we can evaluate whether or not the inductive bias we introduce through FVIN improves prediction accuracy and control performance. ResNN is a competitive baseline, even on lower data regimes ($5$-$10$ trajectories), however it lacks any structure. This makes ResNN likely to over-fit and not capturing long term dynamics.
\subsection{Prediction with Forcing: Generalization and Interpretability}
\begin{figure}[ht!]
    \centering
    \includegraphics[width=0.8\textwidth]{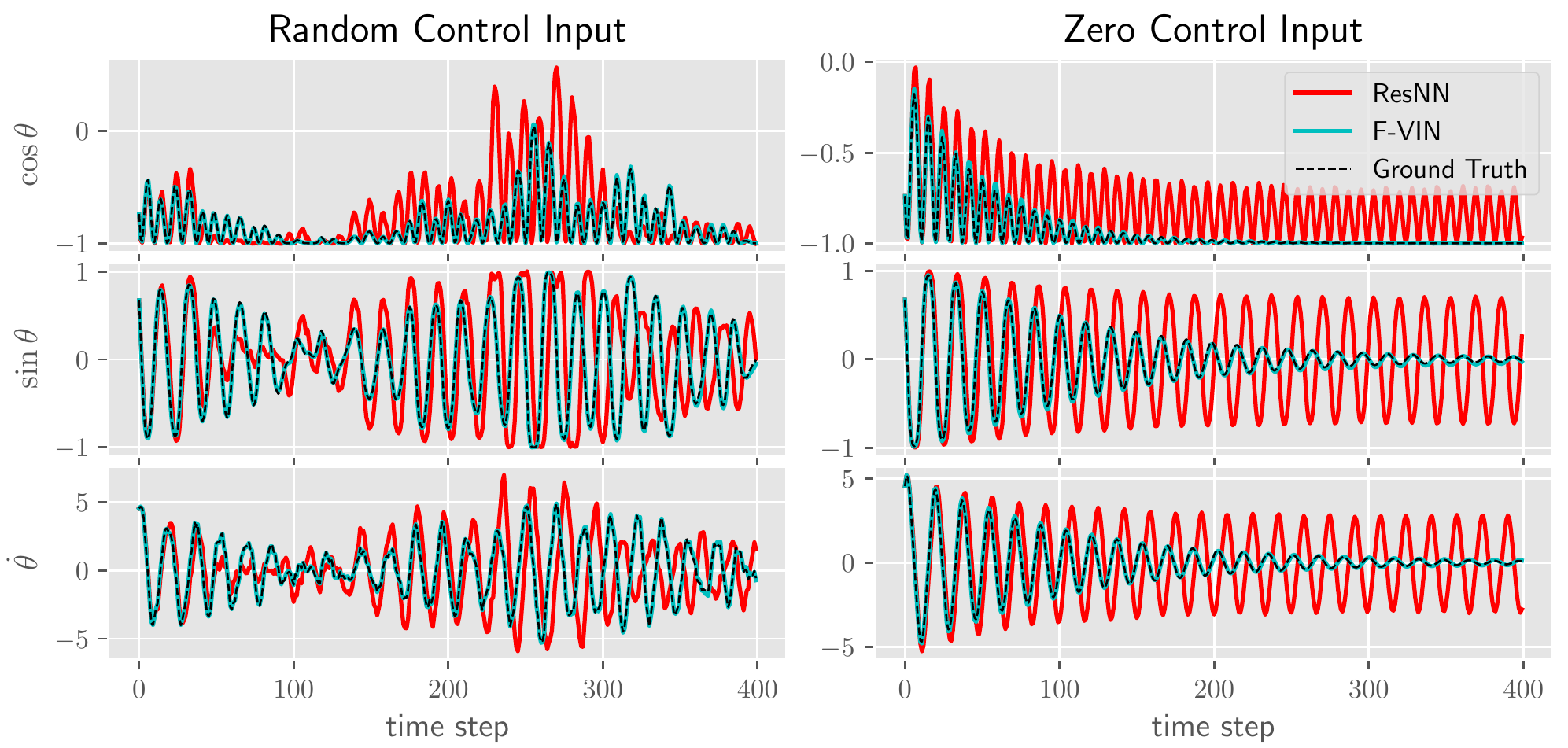}
    \caption{Damped Pendulum: Both VV-FVIN and ResNN models are trained on $5$ trajectories with uniform random control inputs. We compare the prediction performance on a new initial condition with control inputs and without any control. We see that FVIN has better long term prediction error in both settings and captures the true passive dynamics despite having never observed a zero control trajectory. This is an important generalization property of FVIN}
    \label{fig:pend_err}
\end{figure}

\begin{figure}
    \centering
    \includegraphics[width=0.85\textwidth]{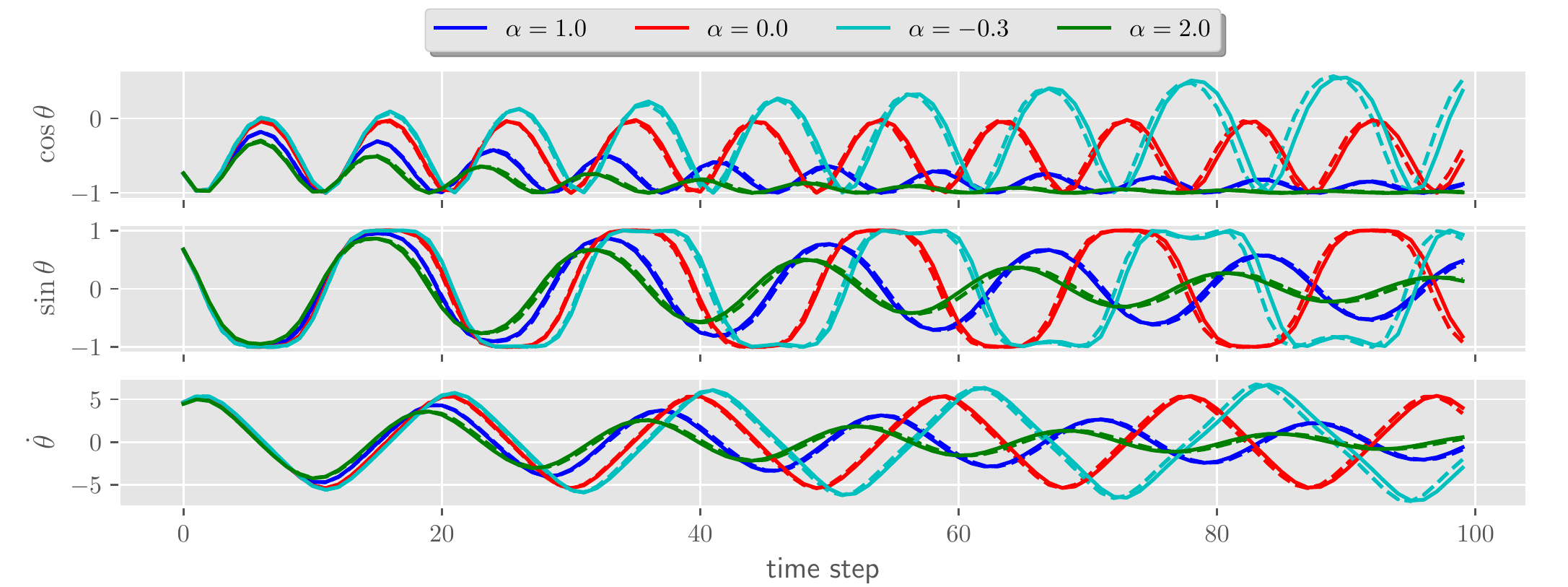}
    \caption{Taking the same learned VV-FVIN from Fig.~\ref{fig:pend_err}, we scale the damping terms in both the FVIN and the pendulum simulation by $\alpha \in \{-0.3, 0, 1.0, 1.5\}$. Dotted lines denote the ground truth generated from simulation. This shows that the model has identified the true damping dissipation and the inductive biased imposed on forcing and passive dynamics are meaningful/interpretable.}
    \label{fig:alpha}
\end{figure}

Although variational integrators are no longer exactly sympletic in the presence of heavy forcing, they still often account for forcing significantly better than other discretizations~\cite{dmoc}. Ideally the FVIN components $F^{(control)}$, $F^{(damping)}$ and $V$ should truly represent the corresponding physical properties of the mechanical system. To study these benefits in the VV-FVIN we consider a damped-pendulum undergoing random control input with the observation states of the form $z=[\cos{\theta}, \sin{\theta}, \dot \theta]$. Only $5$ trajectories, each $50$ time steps and random control inputs, are taken from different initial conditions. After which we test the learned models in two settings: (a) A test trajectory with random control forcing and (b) A test trajectory with no torque inputs, dissipating energy passively.
% \begin{enumerate}%[label=\textbf{\alph*)}]
%     \item A test trajectory with random control forcing.
%     \item A test trajectory with no torque inputs, dissipating energy passively.
% \end{enumerate}
In both cases models produce open-loop trajectories given the same control sequences of the test trajectory. Results for FVIN and ResNN are shown in Fig.~\ref{fig:pend_err} and show that both models are able to track accurately up until the trained horizon length where ResNN starts to diverge. Although the ResNN tracks for a significant period of time, the FVIN almost exactly accounts for forcing due to its variational integrator structure.
More interestingly in Fig.~\ref{fig:pend_err}, we find that even though the models have never been trained on a trajectory without forcing, that under no control inputs the FVIN still tracks very closely to the true passive dynamics of the pendulum where the ResNN captures a very different dissipation behavior. This tells us that the FVIN is actually accounting for forcing due to control allowing it to extract the true passive dynamics from very few trajectories.
In addition to these two cases, we can further investigate the FVIN by removing the additive term $F^{(control)}$ and comparing against the same pendulum simulation without damping. We can even scale the respective damping terms with the parameter $\alpha$  which demonstrates that our model is interpretable with respect to our true mechanical system. Note that we simply use the same FVIN model learned from the same $5$ forced pendulum trajectories with damping. In Fig.~\ref{fig:alpha} we test both positive and negative values of $\alpha$ where $\alpha=0.0$ corresponds to zero damping dissipation.
\subsection{Model-based Control with Forced-VIN}
In the previous section we have shown that FVIN can very accurately predict non-conservative systems with control even with only a few trajectories. In principle, this prediction performance combined with a model-based planning method should translate to superior control performance with respect to a cost function. We aim to emphasize the benefit of accurately accounting for passive dynamics and forcing in a model-predictive control (MPC) setting. Namely, we utilize a general sample-based zeroth-order optimization called the cross-entropy method (CEM)~\cite{rubinstein1997optimization} to compute the optimal control at each time step. CEM uses the learned model to sample and evaluate trajectories and then fits  a multivariate Gaussian to the predicted optimal control sequence. In the MPC setting the CEM optimization is performed at every time step and only the first optimal control in the sequence is applied. More details about CEM and the iterative algorithm used can be found in Appendix~\ref{app:cem}.
We consider two control tasks: (a) A $1$-degree of freedom damped pendulum swing up and stabilize task, (b) A $2$-degree of freedom damped cartpole swing up and stabilize task. Additionally we compare three increasing data regimes for trajectories of length $50$. We expect that both models will learn similar control policies in the highest data regime, however the FVIN should benefit from its inductive biases in the lower data settings. The pendulum trajectories are collected with random sampling and cartpole trajectories use additional policy refinement trajectories for exploration. 
% (a) a swing-up and stabilize task on the damped pendulum system, (b) a stabilizing task to a set point $x=1$ on a damped mass-spring system.
\begin{figure}[ht!]
    \centering
    \includegraphics[width=0.8\textwidth]{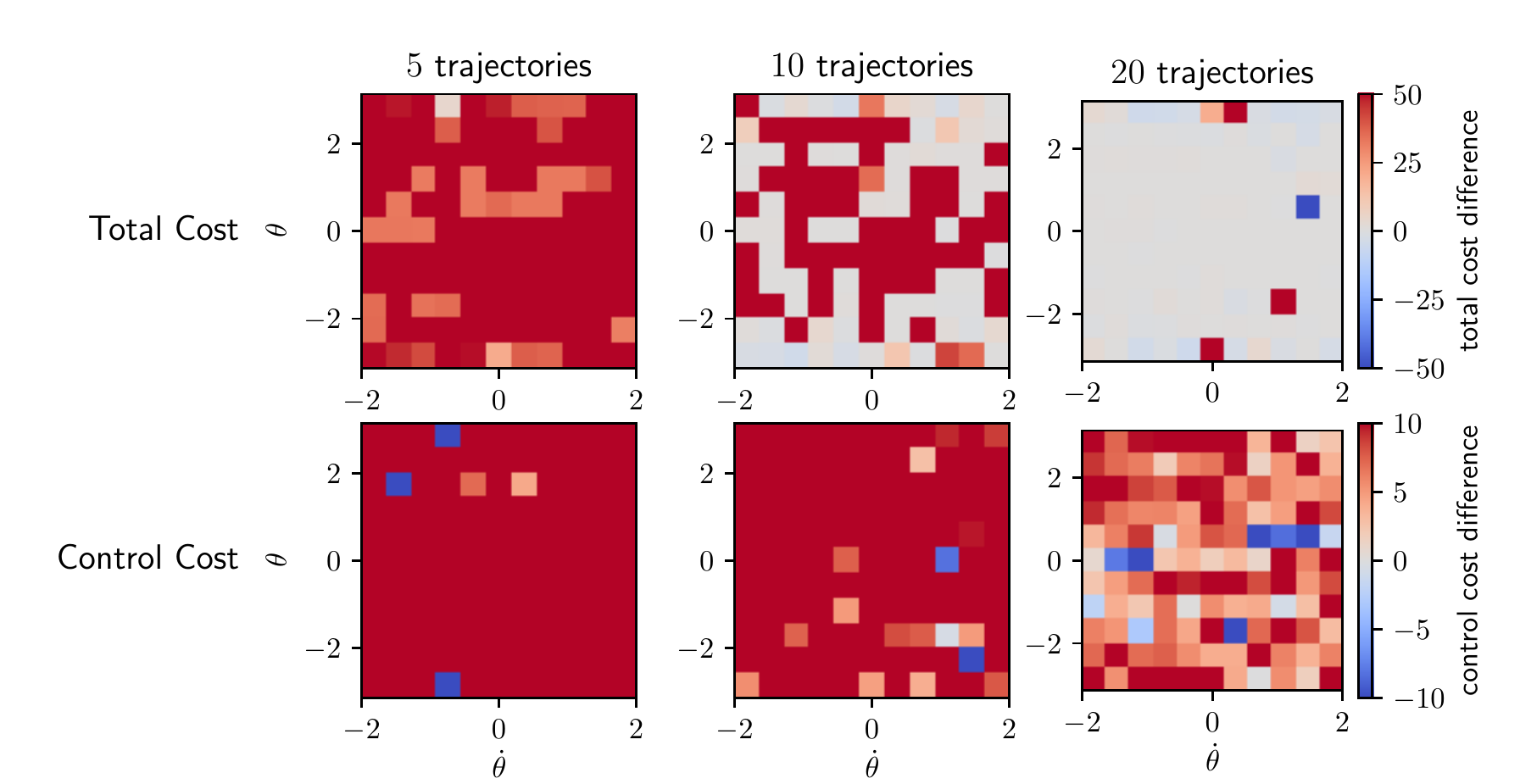}
    \caption{Damped pendulum: We compare MPC total cost (top row) and control effort (bottom row) for VV-FVIN and ResNN models on three data regimes. Each square is an initial condition and color indicates the cost difference (higher red values are in favor of FVIN).}
    \label{fig:mpc_compare}
\end{figure}
\begin{table}[]
\centering
{\renewcommand{\arraystretch}{1.2}
\begin{tabular}{cccc|ccc}
\hline
      & \multicolumn{3}{c|}{Damped Pendulum Swing-up} & \multicolumn{3}{c}{Damped Cart-Pole Swing-up} \\ \hline
      & 5 traj.        & 10 traj.      & 20 traj.     & 20 traj.          & 30 traj.         & 50 traj.         \\ \hline
FVIN success \% & \textbf{1.0}      & \textbf{1.0}     & \textbf{1.0}     & \textbf{0.88}         & \textbf{0.96}          & \textbf{1.0}          \\\hline
ResNN success \%  & 0.12    & 0.74   & 0.99   & 0.02          &  0.43         & 0.88          \\\hline 
\end{tabular}}
\caption{We compare the percentage of successful attempts of the MPC tasks out of $100$ sampled initial conditions using a planning horizon of $H=15$. Success is indicated by achieving a state within an $\epsilon=0.1$ ball of zero position and velocity of the pole.}
%. When FVIN is being used as the model, the task is successful even in low data regimes.}
\label{table:success_rate}
\end{table}
In Fig.~\ref{fig:mpc_compare} we plot the cost difference between FVIN and ResNN MPC on the pendulum system over a grid of initial conditions (higher is better). In Table~\ref{table:success_rate} we see that FVIN is able to successfully solve the damped pendulum at all initial conditions for all data regimes, where ResNN struggles to stabilize at all at in $5$ trajectories and about $74\%$ of the initial conditions at $10$. In the $20$ trajectory setting both models perform well at almost all initial conditions with similar resulting cost. However, according to Fig.~\ref{fig:mpc_compare} the control effort for achieving the same total cost is still significantly less using FVIN. We also compare on the damped cartpole task with similar results, but with $2$-degrees of freedom it requires a greater amount of data.
% \begin{figure}[ht!]
%     \centering
%     \includegraphics[width=\textwidth]{figs/spring_mpc.pdf}
%     \caption{MPC SpringMass}
%     \label{fig:mpc_compare}
% \end{figure}
The FVIN structure enables the MPC controller to take advantage of the underlying passive dynamics even in low-data regimes, making it more likely to produce a high-performing control policy.
\subsection{Evaluation of Real Quanser Pendulum Platform}
\begin{wrapfigure}{R}{0.3\textwidth}
  \begin{center}
    \includegraphics[width=0.5\linewidth]{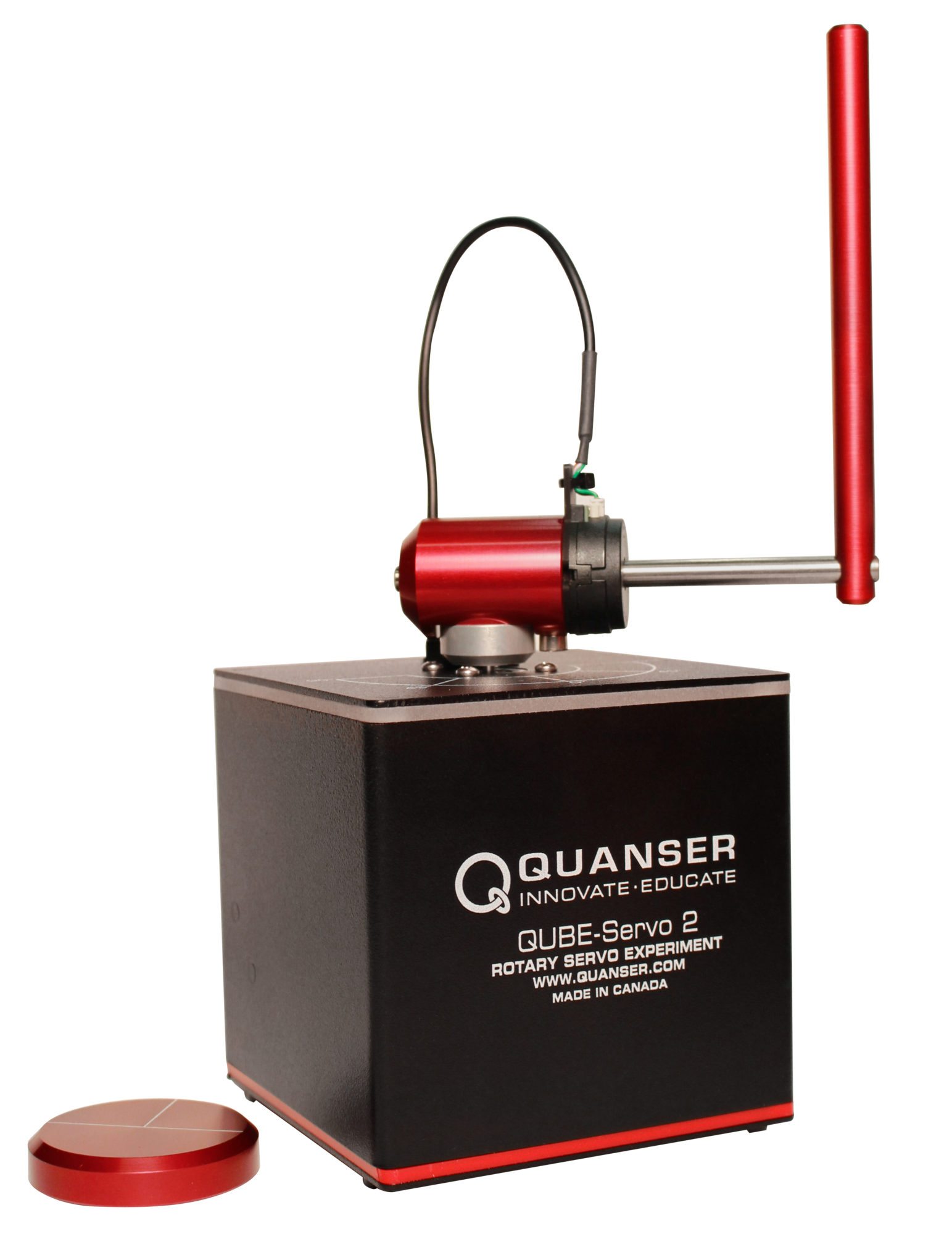}
  \end{center}
  \caption{The Quanser-Qube Servo $2$~\cite{quancube} uses a reinforcement learning-like interface developed by~\citet{polzounov2020blue}.}
  \label{fig:qqs2}
\end{wrapfigure}
\begin{figure}
    \centering
    \includegraphics[width=.95\textwidth]{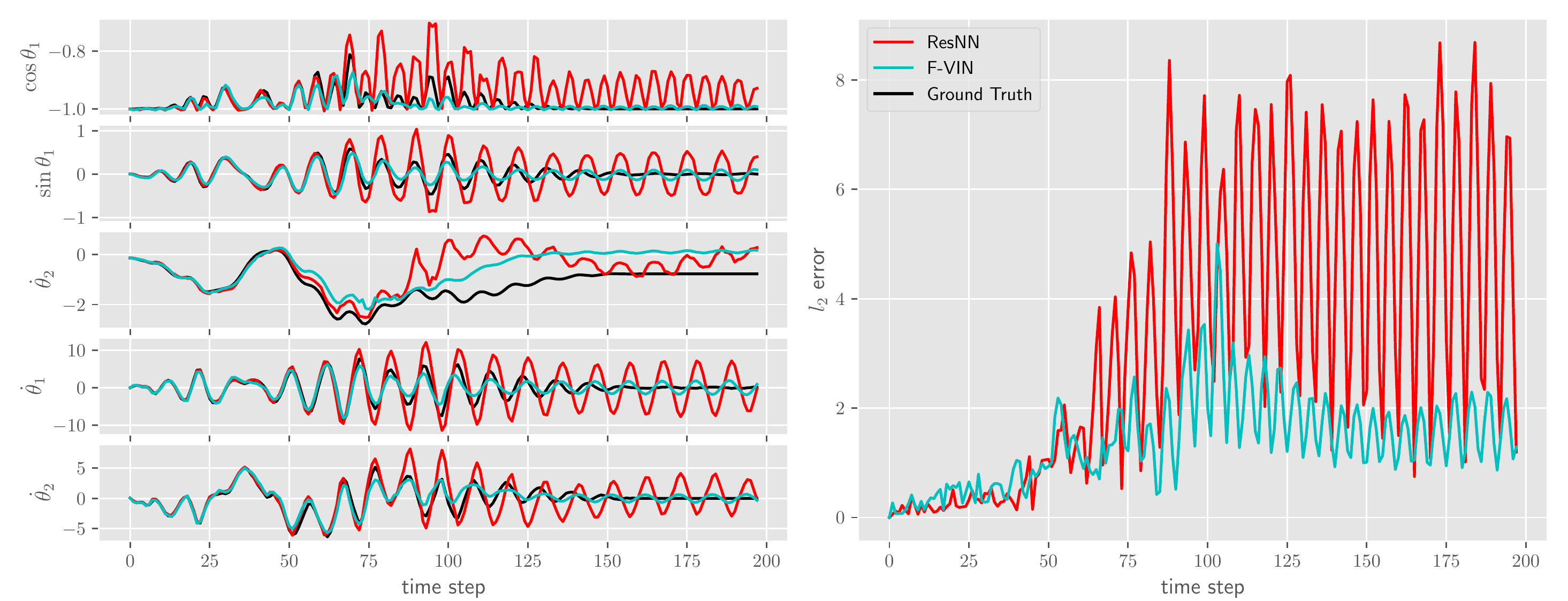}
    \caption{Prediction results on the real QQS$2$ (depicted in Fig.~\ref{fig:qqs2}) from $10$ $50$ step trajectories with random control. In this test trajectory control is turned off after $100$ steps and the dissipation behavior is predicted successfully by the VV-FVIN where ResNN over-fits the first $50$ steps. Note that $\theta_2$ the rotational base does not necessarily go to $0$.}
    \label{fig:real_traj}
\end{figure}

To further provide justification of why the FVIN is a viable choice when modeling mechanical system, we test the prediction performance on measurements from a real system. We use the Quanser-Qube Servo $2$ (QQS2) as seen in Fig.~\ref{fig:qqs2} which consist of $2$ degrees of freedom and $1$ actuation at the base. The QQS$2$ is very similar to a ``cart-pole'' system, but with rotational motion at the base rather than translation. The pendulum angle is given by $\theta_1$ and the base angle $\theta_2$. The QQS$2$ system provides both of these measurements and their velocities via optical encoders. The observation is given as $(\cos{\theta_1}, \sin{\theta_1}, \theta_2, \dot \theta_1, \dot \theta_2)$, so that the autoencoder is necessary to infer the configuration state. We run similar prediction experiments where each model is provided $10$ trajectories of $50$ steps each using random control inputs. We then test on a new trajectory of length $200$ where controls are turned off after step $100$. The idea is that the model should predict the dissipation behavior. In Fig.~\ref{fig:real_traj} we show that although the ResNN accurately predicts accurately for the first $50$ steps, it does not capture any dissipation and has over-fit the training data. On the contrary, FVIN has lower accuracy in the short terms, but is able to predict well past the $50$ step horizon and predict dissipation due to damping in the QQS$2$. 
%===============================================================================
\section{Conclusion}
\label{sec:conclusion}
In this paper we introduce a necessary extension of Variational Integrators Networks~(\cite{pmlr-v108-saemundsson20a}) to enable applications to general non-conservative and controlled systems using the modified variational \textit{Discrete Lagrange-d'Alembert} principle. We then parameterized these new forcing terms in a way where the the non-conservative passive dynamics and forcing could be accurately learned simultaneously from random trajectories. We have shown that these forcing terms can be interpretable by generalizing the learned terms to other levels of damping forcing using the same model. The FVIN is highly data-efficient compared to the baseline ResNN which makes it ideal for model-based control, consistently obtaining more efficient control even in high-data regimes. The FVIN is also capable of predicting on a real non-conservative cartpole and proves to capture energy behavior better than the baseline. 
Going forward we would like to scale this model to larger and more complicated systems and apply it to real-time control applications. Additionally we think it would be important to combine FVIN with the rich field of discrete mechanics and optimal control (DMOC)~(\cite{dmoc}) and implicit definitions for better end-to-end learning control. FVIN may serve as a building block for future efforts in this area.
\acks{This work is supported by the NSF/USDA National AI Institute AIFARMS, AFRI grant no. 2020-67021-32799/project accession no.1024178
9:48
and by USDA/NSF NRI 2.0 project USDA\#2019-67021-28989/NSF\#1830343.}
%===============================================================================
% The maximum paper length is 8 pages excluding references and acknowledgements, and 10 pages including references and acknowledgements
\clearpage
% The acknowledgments are automatically included only in the final version of the paper.
% \acks{If a paper is accepted, the final camera-ready version will (and probably should) include acknowledgments. All acknowledgments go at the end of the paper, including thanks to reviewers who gave useful comments, to colleagues who contributed to the ideas, and to funding agencies and corporate sponsors that provided financial support.}

%===============================================================================

\bibliography{l4dc2021-sample}  % .bib

\begin{thebibliography}{21}
\providecommand{\natexlab}[1]{#1}
\providecommand{\url}[1]{\texttt{#1}}
\expandafter\ifx\csname urlstyle\endcsname\relax
  \providecommand{\doi}[1]{doi: #1}\else
  \providecommand{\doi}{doi: \begingroup \urlstyle{rm}\Url}\fi

\bibitem[Chen et~al.(2018)Chen, Rubanova, Bettencourt, and
  Duvenaud]{neural_ode}
Ricky T.~Q. Chen, Yulia Rubanova, Jesse Bettencourt, and David~K Duvenaud.
\newblock Neural ordinary differential equations.
\newblock In S.~Bengio, H.~Wallach, H.~Larochelle, K.~Grauman, N.~Cesa-Bianchi,
  and R.~Garnett, editors, \emph{Advances in Neural Information Processing
  Systems 31}, pages 6571--6583. Curran Associates, Inc., 2018.

\bibitem[Dullerud and Paganini(2013)]{dullerud2013course}
Geir~E Dullerud and Fernando Paganini.
\newblock \emph{A course in robust control theory: a convex approach},
  volume~36.
\newblock Springer Science \& Business Media, 2013.

\bibitem[Greydanus et~al.(2019)Greydanus, Dzamba, and Yosinski]{hamiltonianNN}
Samuel Greydanus, Misko Dzamba, and Jason Yosinski.
\newblock Hamiltonian neural networks.
\newblock In H.~Wallach, H.~Larochelle, A.~Beygelzimer, F.~d~Alch\'{e}-Buc,
  E.~Fox, and R.~Garnett, editors, \emph{Advances in Neural Information
  Processing Systems 32}, pages 15379--15389. Curran Associates, Inc., 2019.

\bibitem[Guiochet et~al.(2017)Guiochet, Machin, and
  Waeselynck]{guiochet2017safety}
J{\'e}r{\'e}mie Guiochet, Mathilde Machin, and H{\'e}l{\`e}ne Waeselynck.
\newblock Safety-critical advanced robots: A survey.
\newblock \emph{Robotics and Autonomous Systems}, 94:\penalty0 43--52, 2017.

\bibitem[Hochlehnert et~al.(2021)Hochlehnert, Terenin, S{\ae}mundsson, and
  Deisenroth]{hochlehnert2021learning}
Andreas Hochlehnert, Alexander Terenin, Steind{\'o}r S{\ae}mundsson, and Marc
  Deisenroth.
\newblock Learning contact dynamics using physically structured neural
  networks.
\newblock In \emph{International Conference on Artificial Intelligence and
  Statistics}, pages 2152--2160. PMLR, 2021.

\bibitem[Janner et~al.(2019)Janner, Fu, Zhang, and Levine]{model_bias}
Michael Janner, Justin Fu, Marvin Zhang, and Sergey Levine.
\newblock When to trust your model: Model-based policy optimization.
\newblock In H.~Wallach, H.~Larochelle, A.~Beygelzimer, F.~d~Alch\'{e}-Buc,
  E.~Fox, and R.~Garnett, editors, \emph{Advances in Neural Information
  Processing Systems 32}, pages 12519--12530. Curran Associates, Inc., 2019.

\bibitem[Junge et~al.(2005)Junge, Marsden, and
  Ober-Bl{\"o}baum]{junge2005discrete}
Oliver Junge, Jerrold~E Marsden, and Sina Ober-Bl{\"o}baum.
\newblock Discrete mechanics and optimal control.
\newblock \emph{IFAC Proceedings Volumes}, 38\penalty0 (1):\penalty0 538--543,
  2005.

\bibitem[Kingma and Ba(2019)]{kingma2019method}
Diederik~P Kingma and J~Adam Ba.
\newblock A method for stochastic optimization. arxiv 2014.
\newblock \emph{arXiv preprint arXiv:1412.6980}, 434, 2019.

\bibitem[Kingma and Welling(2013)]{kingma2013auto}
Diederik~P Kingma and Max Welling.
\newblock Auto-encoding variational bayes.
\newblock \emph{arXiv preprint arXiv:1312.6114}, 2013.

\bibitem[Krizhevsky et~al.(2012)Krizhevsky, Sutskever, and Hinton]{imagenet}
Alex Krizhevsky, Ilya Sutskever, and Geoffrey~E Hinton.
\newblock Imagenet classification with deep convolutional neural networks.
\newblock In F.~Pereira, C.~J.~C. Burges, L.~Bottou, and K.~Q. Weinberger,
  editors, \emph{Advances in Neural Information Processing Systems 25}, pages
  1097--1105. Curran Associates, Inc., 2012.

\bibitem[Lutter et~al.(2019)Lutter, Ritter, and Peters]{LutterRP19}
Michael Lutter, Christian Ritter, and Jan Peters.
\newblock Deep lagrangian networks: Using physics as model prior for deep
  learning.
\newblock In \emph{7th International Conference on Learning Representations,
  {ICLR} 2019, New Orleans, LA, USA, May 6-9, 2019}. OpenReview.net, 2019.

\bibitem[Marsden and West(2001)]{west}
J.~E. Marsden and M.~West.
\newblock Discrete mechanics and variational integrators.
\newblock \emph{Acta Numerica}, 10:\penalty0 357–514, 2001.
\newblock \doi{10.1017/S096249290100006X}.

\bibitem[Ober-Blöbaum et~al.(2010)Ober-Blöbaum, Junge, and Marsden]{dmoc}
Sina Ober-Blöbaum, Oliver Junge, and Jerrold~E. Marsden.
\newblock Discrete mechanics and optimal control: An analysis.
\newblock \emph{ESAIM: Control, Optimisation and Calculus of Variations},
  17\penalty0 (2):\penalty0 322–352, Mar 2010.
\newblock ISSN 1262-3377.
\newblock \doi{10.1051/cocv/2010012}.

\bibitem[Polzounov et~al.(2020)Polzounov, Sundar, and
  Redden]{polzounov2020blue}
Kirill Polzounov, Ramitha Sundar, and Lee Redden.
\newblock Blue river controls: A toolkit for reinforcement learning control
  systems on hardware, 2020.

\bibitem[{Quanser}(2020)]{quancube}
{Quanser}.
\newblock Quanser-cube servo 2, 2020.
\newblock [Online; accessed July 25, 2020].

\bibitem[Rubinstein(1997)]{rubinstein1997optimization}
Reuven~Y Rubinstein.
\newblock Optimization of computer simulation models with rare events.
\newblock \emph{European Journal of Operational Research}, 99\penalty0
  (1):\penalty0 89--112, 1997.

\bibitem[Saemundsson et~al.(2020)Saemundsson, Terenin, Hofmann, and
  Deisenroth]{pmlr-v108-saemundsson20a}
Steindor Saemundsson, Alexander Terenin, Katja Hofmann, and Marc Deisenroth.
\newblock Variational integrator networks for physically structured embeddings.
\newblock In Silvia Chiappa and Roberto Calandra, editors, \emph{Proceedings of
  the Twenty Third International Conference on Artificial Intelligence and
  Statistics}, volume 108 of \emph{Proceedings of Machine Learning Research},
  pages 3078--3087, Online, 26--28 Aug 2020. PMLR.

\bibitem[Silver et~al.(2018)Silver, Hubert, Schrittwieser, Antonoglou, Lai,
  Guez, Lanctot, Sifre, Kumaran, Graepel, Lillicrap, Simonyan, and
  Hassabis]{Silver1140}
David Silver, Thomas Hubert, Julian Schrittwieser, Ioannis Antonoglou, Matthew
  Lai, Arthur Guez, Marc Lanctot, Laurent Sifre, Dharshan Kumaran, Thore
  Graepel, Timothy Lillicrap, Karen Simonyan, and Demis Hassabis.
\newblock A general reinforcement learning algorithm that masters chess, shogi,
  and go through self-play.
\newblock \emph{Science}, 362\penalty0 (6419):\penalty0 1140--1144, 2018.
\newblock ISSN 0036-8075.
\newblock \doi{10.1126/science.aar6404}.

\bibitem[Verlet(1967)]{verlet}
Loup Verlet.
\newblock Computer "experiments" on classical fluids. i. thermodynamical
  properties of lennard-jones molecules.
\newblock \emph{Phys. Rev.}, 159:\penalty0 98--103, Jul 1967.
\newblock \doi{10.1103/PhysRev.159.98}.

\bibitem[Vujanovic(1978)]{vujanovic1978conservation}
B~Vujanovic.
\newblock Conservation laws of dynamical systems via d'alembert's principle.
\newblock \emph{International Journal of Non-Linear Mechanics}, 13\penalty0
  (3):\penalty0 185--197, 1978.

\bibitem[Wendlandt and Marsden(1997)]{wendlandt1997mechanical}
Jeffrey~M Wendlandt and Jerrold~E Marsden.
\newblock Mechanical integrators derived from a discrete variational principle.
\newblock \emph{Physica D: Nonlinear Phenomena}, 106\penalty0 (3-4):\penalty0
  223--246, 1997.

\end{thebibliography}
\clearpage
\appendix
\section{Architecture and Learning Specifications}
We use quite standard neural network architectures for all components. We use the Adam optimizer~\cite{kingma2019method} with a fixed learning rate of $5\mathrm{e}{-4}$ and batch size of $2048$ is used for all training runs. 
\subsection{Dynamic Prediction}
For the FVIN there are three separate network heads: $V_{\theta}(q_k)$ is for the gradient of the potential-based energy dynamics, $F^{(control)}_{\phi_1}(q_k, u_k)$ is for control forcing terms, and $F^{(\phi_2)})(q_k, \dot q_k)$ is for passive non-conservative terms. The outputs are all added in the form of the VV-FVIN~(\ref{eq:verlet}) to produce the next state. Each head is a $2$-layer fully-connected neural network with $100$ hidden units and ReLU activations. 

The ResNN consist of only two heads, one for position-velocity-dependent dynamics and another dependent on control, position and velocity which are added to produce the residual of the next state. Again, each head is a two-layer fully-connected network with ReLU activations. We found that the ResNN performance does not depend much on the affine separation of forcing and residual dynamics or additional network heads.
\subsection{Encoder-Decoder}
Both models use the same encoder-decoder architectures also consisting of $2$-layer fully-connected networks with $100$ hidden units and ReLU activations. For the case of the VV-FVIN, only the position states are encoded and decoded, where the velocity terms are passed under the identity. This is because the velocities are given in the desired form of the configuration state space. This approach is applied to both FVIN and the baseline. It is possible to also encode velocity states if they are not already in the desired coordinates or perhaps require denoising, however this design choice was avoided for simplicity.
\section{System Details}
The experiments used for this paper consist of two simulation environments and one real system. We will now provide additional details for each system. Each simulation is described by a $2$nd-order ODE and is integrated using the adaptive $4$th order RK45 method with a time step of $t=0.1$.
\subsection{Damped Pendulum}
The damped pendulum consist of a single rotational degree of freedom with damping dissipation dependent on only velocity $\dot \theta$. Control is introduced as a torque $\tau \in [-2,2]$. The governing ODE that fully determines the dynamics can be easily derived through the Euler-Lagrange method and is given by:
\begin{equation}
    \Ddot{\theta} = - \frac{\mu}{m} \dot \theta -\frac{g}{l}\sin\theta  + \frac{\tau}{m l^2}
\end{equation}
where $l$ is the length of the pendulum, $m$ is the mass at the end of the length, $g$ is gravity and $\mu$ is a damping coefficient. In the paper we choose the parameters $m=l=1$, $g=9.81$ and $\mu=0.2$. The observation is given to the model in the form: $(\cos \theta, \sin \theta, \dot \theta)$. A maximum of $50$ steps is used for each trajectory.
\subsection{Damped Cartpole}
The cartpole has one rotational and one translational degree of freedom. There is force on the translational cart $F \in [-10,10]$ and, in addition, damping at both degrees of freedom. The cartpole ODE can be given in the following matrix form:
\begin{equation}
    \begin{bmatrix}
    m_c + m_p & m_p l \cos \theta\\ m_p l \cos \theta & m_p l^2 
    \end{bmatrix}
    \begin{bmatrix}
    \Ddot{x}\\
    \Ddot{\theta}
    \end{bmatrix}
    - 
    \begin{bmatrix}
    m_p l \dot \theta^2 \sin \theta\\
    m_p g l \sin \theta
    \end{bmatrix}
    = \begin{bmatrix}
    F - \mu_c \dot x\\
    -\mu_p \dot \theta
    \end{bmatrix}
\end{equation}
where $m_c$ is the cart mass, $m_p$ the pole mass, $\mu_c$ cart damping coefficient and $\mu_p$ is the pole rotation damping coefficient. In the paper we use $m_c = 1,\, m_p = 0.1,\, l=1, \mu_c=0.1,\, \mu_p=0.05$ and $g=9.81$. The observation finally given to the model is: $(x, \cos \theta, \sin \theta, \dot x, \dot \theta)$. A maximum of $50$ steps is used for each trajectory.
\subsection{Quansar-Qube Servo 2}
The QQS2 system is a real test platform~\cite{quancube} with two rotational degrees of freedom. It mimics a cartpole except for the cart rotates and is driven by torque about the base rather than sliding. The measurements provided by the QQS2 are given via optical encoder of both angles $\theta_1$ and $\theta_2$ and their velocities at $250$ Hz. However we choose to sub-sample these measurements at $25$ Hz ($dt=0.04$), holding the control command constant over the entire period. The measurements given to the model are in the form $(\cos \theta_1, \sin \theta_1,\theta_2, \dot \theta_1,\dot \theta_2)$ and torques are defined by voltage in the range $\tau \in [-2,2]$. Here, $\theta_1$ is the angle of the pole and $\theta_2$ is the angle of the ``cart'' base. For all experiments on the QQS2 we use the interface provided by~\citet{polzounov2020blue}. We sample $50$ steps for each training trajectory and $200$ for test trajectories to predict on long term energy dissipation.
\section{Control Experiment Description}\label{app:cem}
In this paper we present control tasks where a sample-based MPC algorithm (CEM MPC) uses F-VIN and the baseline ResNN as a model to plan. This shows the efficacy of F-VIN for, not only prediction generalization, but also improving control performance. In this section we will go into further detail on CEM and how model learning is implemented.
\subsection{CEM MPC}
CEM is a sample-based zeroth order method, in that it does not require any gradient information. CEM uses a population of control sequences $\{u\}^{(k)}_{t:t+H-1}$ to estimate the sequence which maximizes the given reward function under the dynamic model transition. In a way, we are using our model as a simulator to optimize a trajectory. In this case it is assumed control input is modeled as a vector of independent Gaussians $\{u\}^{(k)}_{t:t+H} \sim \mathcal{N}(\mu_{t:t+H}, \sigma^2_{t:t+H} I)$. The control sequence is initialized to be zero mean and unit variance denoted by $p$. At each MPC planning time step, we sample $M$ trajectories from $q$ and evaluate each trajectory using the forward model and reward function. Finally, we refit $q$ as a new Gaussian distribution over the top $K$ trajectories with the highest reward. After $N$ iterations, we return the first control in the sequence and plan again. The algorithm is described as pseudo code below.
\begin{algorithm}[H]
\caption{CEM MPC}\label{alg:offlinempc}
\begin{algorithmic}
\STATE Inputs: $x_0$, horizon $H$, iterations $N$, samples $M$, number of top performing samples $K$, models $(\Phi, f^{(encoder)}, f^{(decoder)})$ and cost function $C(x,u)$.
\STATE $p \gets \mathcal{N}(0,I)$\;
\FOR{$i=1 : N$}
    \FOR{$k=1 : M$}
    \STATE encode initial states $q_{0} \gets f^{(enc)}(x_0)$
    \STATE $U_k \gets u^{(k)}_{t:t+H-1} \sim p$
    \STATE Run $f$ forward from $q_0$ with $U_k$ to obtain cost $C_k$ with $C$ and $f^{(dec)}$
\STATE $\mathcal{K} \gets \text{argsort}(\{C_i\}_{1:M})_{1:K}$
\STATE Fit mean $\mu_{u_{t:t+H-1}}$ and variance $\sigma_{u_{t:t+H-1}}$ to $U_{k\in \mathcal{K}}$
\STATE $p \gets \mathcal{N}(\mu_{u_{t:t+H-1}}, \sigma_{u_{t:t+H-1})}$
\ENDFOR
\ENDFOR
\STATE \textbf{return} first mean control $u^* \gets \mu_{u_t}$
\end{algorithmic}
\end{algorithm}
Since we use MPC, the CEM routine is performed at every time step and can be quite computational expensive. Although trajectory sampling can be efficiently parallelized to achieve real-time application at frequency above $10$Hz. For now, the CEM MPC operates at about $3$Hz on a laptop CPU.
\subsection{Model Learning Routine}
To perform the entire model learning routine, we first initialized a model with $5$ trajectories with randomly sampled control. The model is trained for $5000$ epochs, after which additional trajectories are collected using the CEM MPC policy with Gaussian noise. After each additional trajectory, we train the same model for $1000$ epochs. For CEM parameters, we choose the number of samples trajectory at each iteration to be $M=1000$ and take the top $K=10$ trajectories to refit the control sequence. We choose to run $N=5$ iterations of CEM at each MPC step.
\begin{algorithm}[H]
\caption{Training with CEM MPC}
\label{alg:overall}
\begin{algorithmic}
\STATE Inputs: planning horizon $H$, number of iterations $N$, models $(\Phi, f^{(encoder)}, f^{(decoder)})$, parameters $\Theta = (\theta, \phi, \psi, \zeta)$ and cost function $C(x,u)$.
\STATE Initialize dataset $D$ with random trajectories and train parameters $\Theta$ with loss function L~\eqref{eq:loss}.
\FOR{$i = 1$ \TO $N$}
    \STATE Run CEM MPC policy with Gaussian noise to collect new data $D_i$.
    \STATE $D \gets D \cup D_i$.
    \STATE Train $\Theta$ using the loss function $L$ and data $D$.
\ENDFOR{}
\end{algorithmic}
\end{algorithm}
At the end of the routine, the model is used with the CEM MPC policy without noise to be evaluated over several initial conditions as presented in Fig.~\ref{fig:mpc_compare} and Table~\ref{table:success_rate}.
\subsection{System Cost Functions}
For the Damped Pendulum system we choose a simple quadratic cost function that primarily penalizes the angle $\theta$ with a relatively lower cost on velocity $\dot \theta$ and control. The angle $\theta$ is obtained by transforming the decoded state $\cos \theta$ and $\sin \theta$ with the $\arctan 2$ function.
\begin{equation}
    C(\theta, \dot \theta, u) = \theta^2 + 0.01 \dot \theta^2 + 0.001 u^2
\end{equation}
For the Damped Cartpole system we choose a similar cost, but now also penalizing the position and velocity of the cart ($x,\dot x$).
\begin{equation}
    C(x,\theta, \dot x, \dot \theta, u) = 5 \theta^2 + x^2 + 0.1 \dot x^2 + 0.1 \dot \theta^2 + 0.01 u^2 
\end{equation}
Of course one does not have to use a quadratic cost and many other functions can be used with CEM. Even soft constraints could be enforced by assigning constraint sets with a very high cost.
\section{Additional Figures and Experiments}
\subsection{Damped Cartpole Prediction}
In addition to the control experiment results presented in Table~\ref{table:success_rate} for the Damped Cartpole, we will also demonstrate prediction performance similarly to how we demonstrated prediction for the Damped Pendulum. We collect $25$ trajectories with random control of length $50$ steps to train the VV-FVIN and ResNN baseline. Both models are then evaluated on test trajectories with random control and with zero control. Similarly to the Damped Pendulum, we see that VV-FVIN is still able to generalize to the passive dynamics and predict the long term energy behavior due to damping. We also see improved performance in control prediction which is why the F-VIN was more successful at solving the Damped Cartpole task in the main paper~(\ref{table:success_rate}).
\begin{figure}[h]
    \centering
    \includegraphics[width=1.0\textwidth]{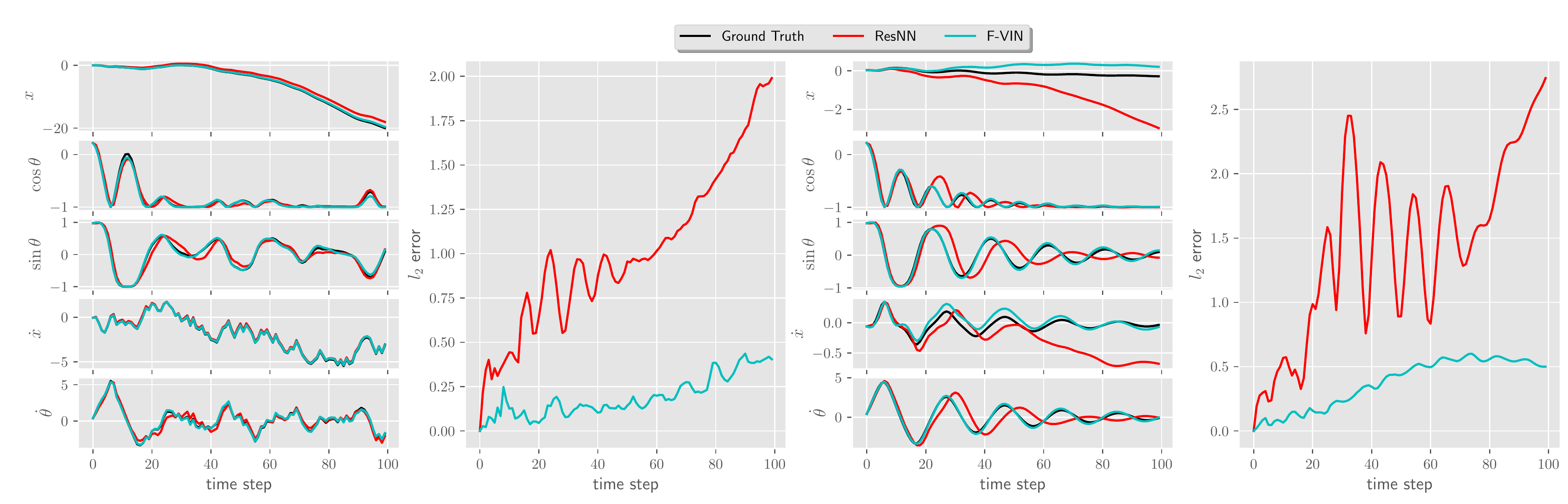}
    \caption{Damped Cartpole: Results for FVIN prediction on the $2$-DOF Damped Cartpole environment. The experiment is reported similarly to Fig.~\ref{fig:pend_err}, except that we use $25$ trajectories, each of length $50$ time steps, to train.} We plot the trajectory and $l_2$ norm error to ground truth for a test trajectory with random control and a test trajectory with zero controls. We use $25$ training trajectories with random controls each of length $50$ steps ($5$ seconds).
    \label{fig:cart_err}
\end{figure}
\subsection{SV-FVIN Prediction}\label{app:sv}
Although in the paper we focus on the position-velocity variant of the FVIN (VV-FVIN), we will also discuss the efficacy of the position-only SV-FVIN given by Eq.~\ref{eq:stormer}. The VV-FVIN and SV-FVIN differ in mostly pragmatic ways. Both integrators use the same trapezoidal approximation of the discrete Lagrangian. We focus on the VV-FVIN mainly because velocity data is often available and the VV-FVIN will provide synchronous estimates of both position and velocity which is more practical for planning and control applications.

To compare fairly against the baseline, we slightly modify the ResNN to rely on both the current and previous state. We show that the SV-FVIN can achieve similar performance and generalize to passive dynamics on the same Damped-Pendulum system restricted to position only measurements. This becomes important in some situations where velocity measurements may not be available. This could be the case if the pendulum position is determined by some remote position sensor or the measurement itself is a raw image. However, it then becomes more difficult to define a control objective on measurements like images. In this paper we focus on more actionable measurements (position-velocity).
\begin{figure}[h]
    \centering
    \includegraphics[width=1.0\textwidth]{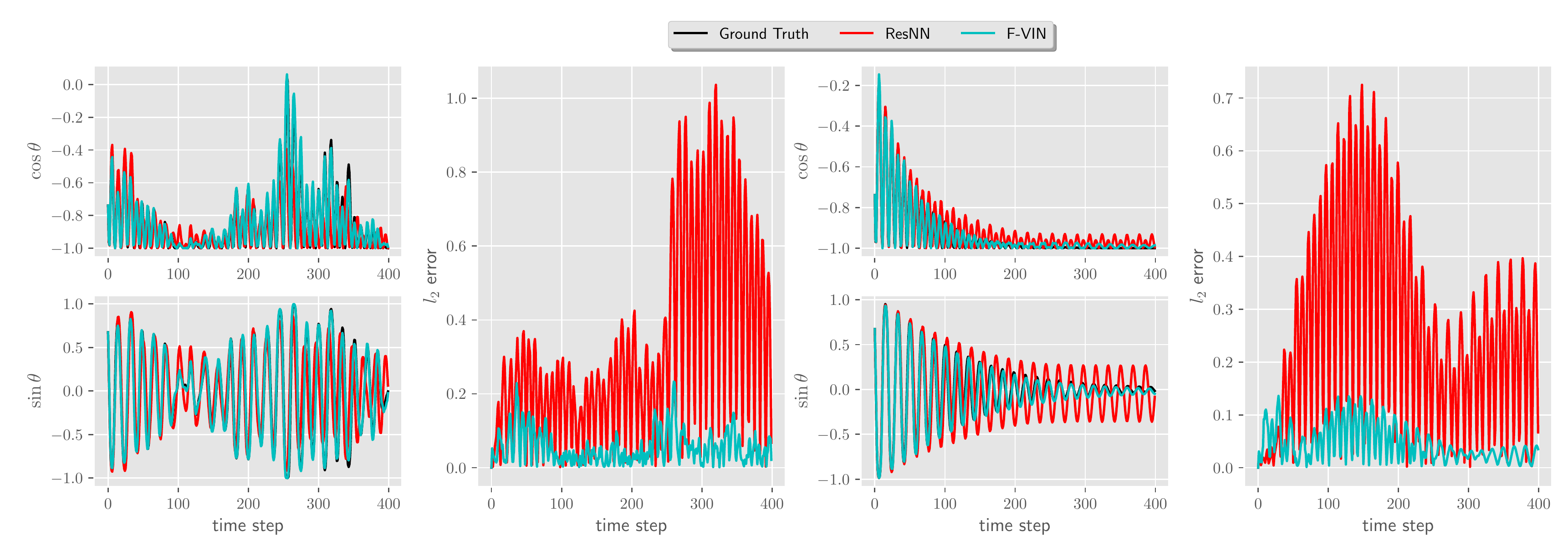}
    \caption{Damped Position-only Pendulum: We plot the trajectory and $l_2$ norm error with respect to the ground truth for a test trajectory with random controls and a test trajectory with zero controls. We use $10$ training trajectories with random controls each of length $50$ steps ($5$ seconds).}
    \label{fig:my_label2}
\end{figure}

% % Acknowledgments---Will not appear in anonymized version
% \acks{We thank a bunch of people.}

%\bibliography{example}

\end{document}